\documentclass[conference]{IEEEtran}
\IEEEoverridecommandlockouts

\usepackage{cite}
\usepackage{amsmath,amssymb,amsfonts}
\usepackage{algorithmic}
\usepackage{graphicx}
\usepackage{textcomp}
\usepackage{xcolor}
\usepackage{multirow}
\usepackage{jabbrv}

\def\BibTeX{{\rm B\kern-.05em{\sc i\kern-.025em b}\kern-.08em
    T\kern-.1667em\lower.7ex\hbox{E}\kern-.125emX}}

\makeatletter 
\newcommand{\linebreakand}{%
\end{@IEEEauthorhalign}
\hfill\mbox{}\par
\mbox{}\hfill\begin{@IEEEauthorhalign}
}
\makeatother 

\begin{document}

\title{Imagined Speech and Visual Imagery as Intuitive Paradigms for Brain-Computer Interfaces\\
\thanks{This work was partly supported by Institute of Information \& Communications Technology Planning \& Evaluation (IITP) grant funded by the Korea government (MSIT) (No. RS--2021--II--212068, Artificial Intelligence Innovation Hub, No. RS--2024--00336673, AI Technology for Interactive Communication of Language Impaired Individuals, and No. RS--2019--II190079, Artificial Intelligence Graduate School Program (Korea University)).}

}

\author{\IEEEauthorblockN{Seo-Hyun Lee}
\IEEEauthorblockA{\textit{Dept. of Brain and Cognitive Engineering} \\
\textit{Korea University} \\
Seoul, Republic of Korea \\
seohyunlee@korea.ac.kr}

\and

\IEEEauthorblockN{Ji-Ha Park}
\IEEEauthorblockA{\textit{Dept. of Artificial Intelligence} \\
\textit{Korea University} \\
Seoul, Republic of Korea \\
jiha\_park@korea.ac.kr}

\and

\IEEEauthorblockN{Deok-Seon Kim}
\IEEEauthorblockA{\textit{Dept. of Artificial Intelligence} \\
\textit{Korea University}\\
Seoul, Republic of Korea \\
deokseon\_kim@korea.ac.kr}
}

\maketitle

\begin{abstract}

Brain-computer interfaces (BCIs) have shown promise in enabling communication for individuals with motor impairments. Recent advancements like brain-to-speech technology aim to reconstruct speech from neural activity. However, decoding communication-related paradigms, such as imagined speech and visual imagery, using non-invasive techniques remains challenging. This study analyzes brain dynamics in these two paradigms by examining neural synchronization and functional connectivity through phase-locking values (PLV) in EEG data from 16 participants. Results show that visual imagery produces higher PLV values in visual cortex, engaging spatial networks, while imagined speech demonstrates consistent synchronization, primarily engaging language-related regions. These findings suggest that imagined speech is suitable for language-driven BCI applications, while visual imagery can complement BCI systems for users with speech impairments. Personalized calibration is crucial for optimizing BCI performance.
\end{abstract}




\section{INTRODUCTION}
The human brain signals contain various information about actions and internal imageries, making it crucial for interpreting intentions and facilitating neural communication. Brain-computer interfaces (BCIs) leverage this potential by translating neural activity into computer commands, allowing individuals to control external devices directly through brain signals. Particularly, BCIs have shown significant promise in assisting those with severe motor impairments, such as individuals with paralysis or locked-in syndrome, enabling them to interact with their environment using only their neural signals \cite{jeong2019classification, chaudhary2016brain, han2020classification}. Recent advancement in this field is brain-to-speech (BTS) technology, which aims to reconstruct audible speech directly from neural activity \cite{moses2021neuroprosthesis, lee2022toward}.

While conventional BCI research focus on decoding motor-related signals or generating speech from spoken or mimed brain signals through invasive methods \cite{anumanchipalli2019speech, lee2020neural, willett2023high, lim2000text}, the challenge of decoding endogeneous and communication-related paradigms, such as imagined speech or visual imagery using non-invasive techniques remains significant \cite{song2017novel, lee2023AAAI}. Imagined speech, which allows individuals to mentally simulate speech without physical articulation, presents a more practical communication mode for those unable to speak \cite{chartier2018encoding, anumanchipalli2019speech}. Visual imagery can also be an effective and intuitive BCI paradigm for communication, as it is not limited by the number of classes and is easy for users to imagine.

Recently, the paradigms of imagined speech and visual imagery have garnered interest in the field of intuitive BCIs, as they directly engage user intention \cite{prabhakar2020framework}. Ongoing efforts focus on robustly decoding these two paradigms while investigating their intrinsic features. However, the underlying features and cortical networks associated with imagined speech and visual imagery remain largely unexplored. A comprehensive understanding of these intrinsic cortical networks may significantly enhance the decoding performance of BCI paradigms \cite{chaudhary2016brain, karikari2023review, lee2020uncertainty}. Previous studies have demonstrated the feasibility of decoding imagined speech at various levels, including phonemes and simple sentences \cite{metzger2023high, kim2018discriminative}, yet multiclass decoding accuracy remains modest, however, visual imagery performance is typically lower \cite{kosmyna2018attending}.


The involvement of specific brain regions, such as Wernicke's area and Broca's area, has been established in the context of imagined speech \cite{moses2021neuroprosthesis}, while studies have shown variations in brain activity during imagery, including decreased activity and high-frequency activation in relevant areas \cite{chaudhary2016brain, karikari2023review, lee2020uncertainty}. Additionally, visual imagery has been linked to the activation of the temporal and occipital regions, utilizing similar neural mechanisms as visual perception.

This study is an expanded version of previous study\cite{lee2021functional}, which investigate the brain dynamics of the imagined speech and visual imagery paradigms by comparing among two paradigms in thirteen distinct classes, with brain connectivity analysis during imagery tasks and resting states. We analyze functional connectivity through phase-locking values (PLV) across specific frequency ranges and various cortical regions. By focusing on the alterations in brain states during the imagery of both paradigms, we aim to identify the key features of brain dynamics that effectively represent imagined speech and visual imagery. Understanding these dynamics could contribute to more precise analyses of EEG data, enhancing the BCI performance \cite{lawhern2018eegnet, kim2023diff}.


\begin{figure}[t]
\centerline{\includegraphics[width=0.99\columnwidth]{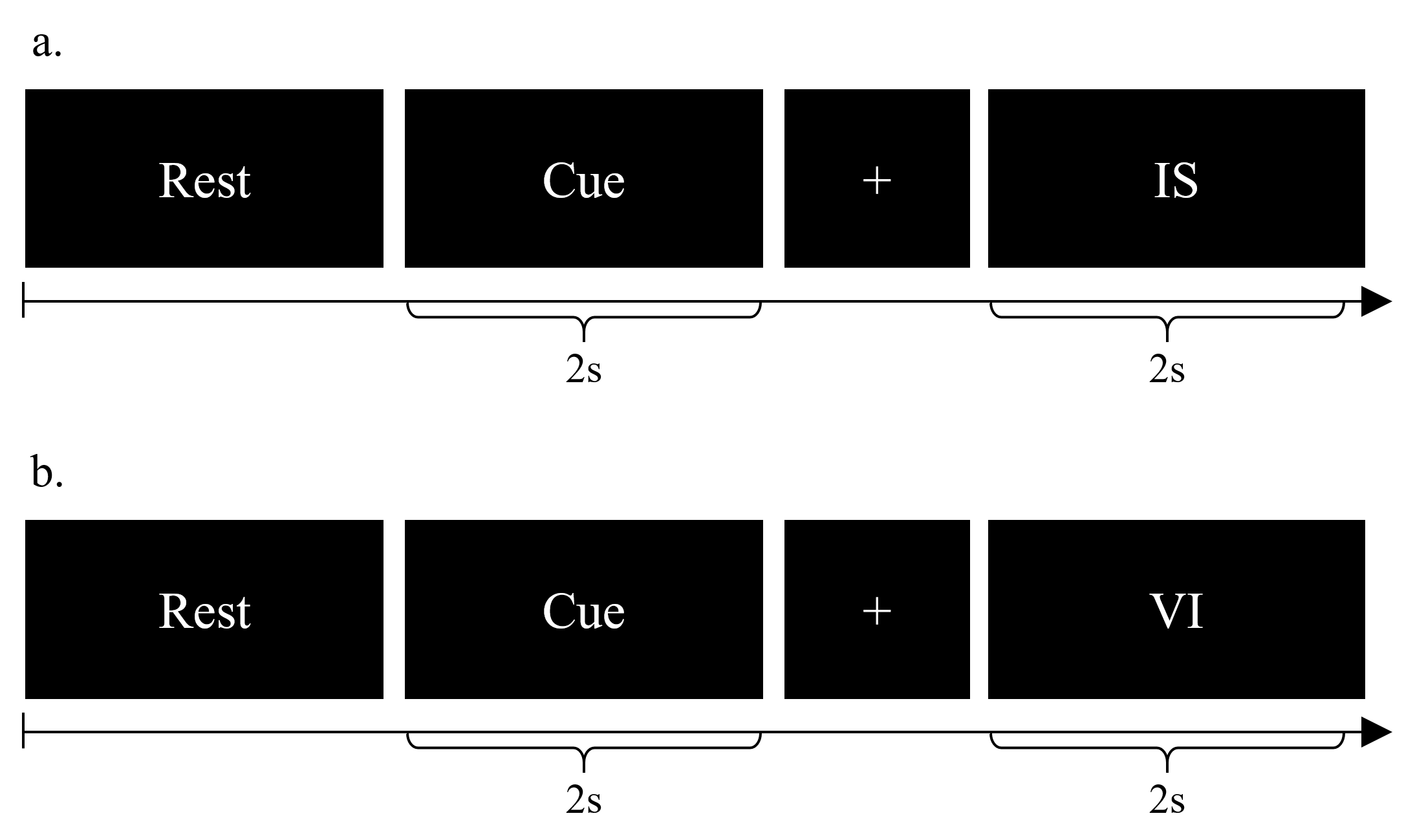}}
\caption{The overall experimental paradigm was designed to investigate the cognitive processes related to (a) imagined speech and (b) visual imagery through specific task performance while recording brain activity. The experimental setup for collecting EEG signals involved placing a EEG cap on participants in a quiet environment to ensure accurate data acquisition.}
\label{fig1}
\end{figure}

\section{METHODS}

\subsection{Dataset Description}
The dataset from the previous work was analyzed in the study \cite{lee2020neural}. This research was conducted in accordance with the Declaration of Helsinki, and informed consent was obtained from sixteen healthy participants. EEG data were collected using a 64-channel cap with active Ag/AgCl electrodes. The FCz channel was used as the reference, and the FPz channel served as the ground. Data acquisition was performed with a Brain Vision Recorder system (BrainProducts GmbH, Germany) and was managed using MATLAB 2018a.






The experimental design consisted of two separate sessions. In session 1, participants engaged in imagined speech, while session 2 involved visual imagery. Either an auditory cue (for imagined speech) or a visual cue (for visual imagery) presented for 2 seconds, followed by a cross mark displayed for a random interval between 0.8 and 1.2 seconds. After the cross mark disappeared, a black screen appeared for another 2 seconds. Participants were instructed to initiate the task as soon as the cross mark was removed. Following 2 seconds of task performance, the cross mark reappeared for 0.8 to 1.2 seconds, succeeded by an additional 2-second black screen (Fig.~\ref{fig1}).

In session 1, participants were asked to imagine saying the given word as though they were speaking, without any physical articulation or sound production. In session 2, they were instructed to visualize the scene corresponding to the presented class. For the rest condition in both sessions, participants remained calm and avoided any intentional brain activity. Clear instructions were provided to ensure that participants did not engage in unrelated cognitive activities during each session, including minimizing body movements and eye blinks while imagining or receiving cues. The cross mark duration was randomized between 0.8 and 1.2 seconds to reduce anticipation regarding task onset.

\begin{table*}[t]
    \centering
    \caption{Average brain connectivity of each class in imagined speech paradigm.} 
    \renewcommand{\arraystretch}{1.08} 
\setlength{\tabcolsep}{7pt}
\resizebox{0.95\textwidth}{!}{
\begin{tabular}{c|cccccccccccc}
\hline
              & \textbf{Ambulance} & \textbf{Clock} & \textbf{Hello} & \textbf{Help me} & \textbf{Light} & \textbf{Pain} & \textbf{Stop} & \textbf{Thank you} & \textbf{Toilet} & \textbf{TV}   & \textbf{Water} & \textbf{Yes}  \\ \hline
\textbf{S1}   & 0.27               & 0.26           & 0.26           & 0.26             & 0.26           & 0.27          & 0.27          & 0.26               & 0.27            & 0.26          & 0.26           & 0.25          \\
\textbf{S2}   & 0.46               & 0.38           & 0.64           & 0.51             & 0.69           & 0.51          & 0.42          & 0.60               & 0.33            & 0.56          & 0.47           & 0.43          \\
\textbf{S3}   & 0.26               & 0.26           & 0.26           & 0.25             & 0.26           & 0.26          & 0.26          & 0.26               & 0.26            & 0.26          & 0.25           & 0.26          \\
\textbf{S4}   & 0.30               & 0.30           & 0.30           & 0.30             & 0.33           & 0.30          & 0.31          & 0.35               & 0.34            & 0.31          & 0.36           & 0.32          \\
\textbf{S5}   & 0.24               & 0.24           & 0.24           & 0.24             & 0.25           & 0.24          & 0.24          & 0.25               & 0.24            & 0.24          & 0.24           & 0.24          \\
\textbf{S6}   & 0.25               & 0.26           & 0.25           & 0.26             & 0.25           & 0.25          & 0.25          & 0.25               & 0.25            & 0.25          & 0.25           & 0.24          \\
\textbf{S7}   & 0.31               & 0.32           & 0.31           & 0.31             & 0.32           & 0.32          & 0.31          & 0.31               & 0.31            & 0.31          & 0.31           & 0.33          \\
\textbf{S8}   & 0.33               & 0.36           & 0.35           & 0.37             & 0.31           & 0.35          & 0.30          & 0.35               & 0.33            & 0.35          & 0.36           & 0.32          \\
\textbf{S9}   & 0.22               & 0.22           & 0.22           & 0.22             & 0.22           & 0.21          & 0.22          & 0.22               & 0.22            & 0.22          & 0.21           & 0.21          \\
\textbf{S10}  & 0.24               & 0.24           & 0.23           & 0.23             & 0.24           & 0.24          & 0.24          & 0.24               & 0.23            & 0.24          & 0.23           & 0.24          \\
\textbf{S11}  & 0.30               & 0.29           & 0.29           & 0.29             & 0.30           & 0.29          & 0.31          & 0.29               & 0.30            & 0.29          & 0.29           & 0.31          \\
\textbf{S12}  & 0.27               & 0.27           & 0.27           & 0.27             & 0.27           & 0.27          & 0.27          & 0.27               & 0.27            & 0.27          & 0.27           & 0.27          \\
\textbf{S13}  & 0.23               & 0.22           & 0.23           & 0.23             & 0.23           & 0.22          & 0.22          & 0.23               & 0.22            & 0.22          & 0.22           & 0.23          \\
\textbf{S14}  & 0.23               & 0.23           & 0.24           & 0.23             & 0.22           & 0.23          & 0.23          & 0.23               & 0.23            & 0.23          & 0.23           & 0.23          \\
\textbf{S15}  & 0.26               & 0.25           & 0.25           & 0.25             & 0.25           & 0.25          & 0.25          & 0.25               & 0.25            & 0.25          & 0.25           & 0.26          \\
\textbf{S16}  & 0.31               & 0.31           & 0.31           & 0.32             & 0.31           & 0.32          & 0.32          & 0.33               & 0.31            & 0.32          & 0.32           & 0.31          \\ \hline
\textbf{Avg.} & \textbf{0.28}      & \textbf{0.28}  & \textbf{0.29}  & \textbf{0.28}    & \textbf{0.29}  & \textbf{0.28} & \textbf{0.28} & \textbf{0.29}      & \textbf{0.27}   & \textbf{0.29} & \textbf{0.28}  & \textbf{0.28} \\ 
\textbf{Std.} & \textbf{0.06}      & \textbf{0.05}  & \textbf{0.10}  & \textbf{0.07}    & \textbf{0.11}  & \textbf{0.07} & \textbf{0.05} & \textbf{0.09}      & \textbf{0.04}   & \textbf{0.08} & \textbf{0.07}  & \textbf{0.06}  \\ \hline
\end{tabular}}
\label{tab1}
\end{table*}

\begin{table*}[t]
    \centering
    \caption{Average brain connectivity of each class in visual imagery paradigm.} 
    \renewcommand{\arraystretch}{1.08} 
\setlength{\tabcolsep}{7pt}
\resizebox{0.95\textwidth}{!}{
\begin{tabular}{c|cccccccccccc}
\hline
              & \textbf{Ambulance} & \textbf{Clock} & \textbf{Hello} & \textbf{Help me} & \textbf{Light} & \textbf{Pain} & \textbf{Stop} & \textbf{Thank you} & \textbf{Toilet} & \textbf{TV}   & \textbf{Water} & \textbf{Yes}  \\ \hline
\textbf{S1}   & 0.50               & 0.53           & 0.51           & 0.53             & 0.52           & 0.51          & 0.51          & 0.52               & 0.52            & 0.52          & 0.51           & 0.51          \\
\textbf{S2}   & 0.24               & 0.24           & 0.24           & 0.24             & 0.23           & 0.24          & 0.24          & 0.24               & 0.23            & 0.25          & 0.23           & 0.24          \\
\textbf{S3}   & 0.21               & 0.22           & 0.22           & 0.22             & 0.22           & 0.23          & 0.22          & 0.22               & 0.23            & 0.22          & 0.22           & 0.22          \\
\textbf{S4}   & 0.37               & 0.35           & 0.35           & 0.37             & 0.35           & 0.37          & 0.34          & 0.34               & 0.34            & 0.37          & 0.35           & 0.36          \\
\textbf{S5}   & 0.25               & 0.25           & 0.26           & 0.25             & 0.26           & 0.26          & 0.26          & 0.26               & 0.25            & 0.25          & 0.26           & 0.25          \\
\textbf{S6}   & 0.23               & 0.22           & 0.22           & 0.29             & 0.23           & 0.26          & 0.23          & 0.22               & 0.22            & 0.22          & 0.21           & 0.26          \\
\textbf{S7}   & 0.42               & 0.51           & 0.51           & 0.60             & 0.47           & 0.60          & 0.51          & 0.38               & 0.42            & 0.60          & 0.35           & 0.46          \\
\textbf{S8}   & 0.26               & 0.28           & 0.28           & 0.28             & 0.26           & 0.28          & 0.26          & 0.28               & 0.27            & 0.28          & 0.28           & 0.26          \\
\textbf{S9}   & 0.24               & 0.23           & 0.23           & 0.24             & 0.23           & 0.23          & 0.24          & 0.23               & 0.23            & 0.23          & 0.23           & 0.23          \\
\textbf{S10}  & 0.24               & 0.24           & 0.24           & 0.23             & 0.24           & 0.24          & 0.24          & 0.24               & 0.24            & 0.24          & 0.24           & 0.25          \\
\textbf{S11}  & 0.28               & 0.27           & 0.27           & 0.27             & 0.28           & 0.27          & 0.28          & 0.27               & 0.28            & 0.27          & 0.27           & 0.27          \\
\textbf{S12}  & 0.24               & 0.25           & 0.25           & 0.24             & 0.24           & 0.25          & 0.24          & 0.25               & 0.24            & 0.24          & 0.25           & 0.24          \\
\textbf{S13}  & 0.24               & 0.24           & 0.25           & 0.24             & 0.24           & 0.24          & 0.25          & 0.24               & 0.24            & 0.24          & 0.24           & 0.25          \\
\textbf{S14}  & 0.25               & 0.24           & 0.25           & 0.24             & 0.25           & 0.24          & 0.25          & 0.24               & 0.25            & 0.25          & 0.24           & 0.25          \\
\textbf{S15}  & 0.26               & 0.26           & 0.25           & 0.26             & 0.27           & 0.26          & 0.27          & 0.26               & 0.26            & 0.26          & 0.27           & 0.27          \\
\textbf{S16}  & 0.29               & 0.27           & 0.30           & 0.29             & 0.29           & 0.29          & 0.30          & 0.30               & 0.30            & 0.30          & 0.30           & 0.28          \\ \hline
\textbf{Avg.} & \textbf{0.28}      & \textbf{0.29}  & \textbf{0.29}  & \textbf{0.30}    & \textbf{0.29}  & \textbf{0.30} & \textbf{0.29} & \textbf{0.28}      & \textbf{0.28}   & \textbf{0.30} & \textbf{0.28}  & \textbf{0.29} \\ 
\textbf{Std.} & \textbf{0.08}      & \textbf{0.10}  & \textbf{0.09}  & \textbf{0.11}    & \textbf{0.09}  & \textbf{0.11} & \textbf{0.09} & \textbf{0.08}      & \textbf{0.08}   & \textbf{0.11} & \textbf{0.07}  & \textbf{0.09} \\ \hline
\end{tabular}}
\label{tab2}
\end{table*}

\subsection{Brain Decoding and Connectivity Analysis}

We analyzed brain connectivity through the PLV measure to understand the brain’s functional changes during imagined speech and visual imagery paradigms. This measure quantifies the synchronization of brain electrical activity, serving as an indicator of functional connectivity. We computed the PLV for both imagery and resting states across four frequency bands and seven cortical regions (including six specific cortical groups as well as an overall group of 64 channels) using the following equations:

\begin{equation}
PLV_{t,i,k}=\displaystyle{\frac{i}{N}|\sum_{n=1}^{N}\exp{(j\theta_{i,k}(t,n))}|}\label{eq}   
\end{equation}

\begin{equation}
\displaystyle{\theta_{i,k}(t,n) = \phi_i(t,n) - \phi_k(t,n);}\label{eq}   
\end{equation}

In these equations, \(\theta_{i,k}(t,n)\) represents the phase difference between channels \(i\) and \(k\) during trial \(n\), and \(N\) stands for the total trial count. Given that PLV captures connectivity between channel pairs, we computed the grand average across all possible channel pairs within each cortical region. 

\begin{table}[t]

\setlength{\tabcolsep}{5pt}
\renewcommand{\arraystretch}{1.04}
\caption{Brain connectivity for imagined speech and visual imagery across different cortical regions.}
\begin{center}

\begin{tabular}{c|cccc|cccc}
\hline
\multicolumn{1}{l|}{{\color[HTML]{333333} \textbf{}}} & {\color[HTML]{333333} IS}            & {\color[HTML]{333333} Rest}          & {\color[HTML]{333333} t}               & {\color[HTML]{333333} p}              & {\color[HTML]{333333} VI}            & {\color[HTML]{333333} Rest}          & {\color[HTML]{333333} t}               & {\color[HTML]{333333} p}              \\ \hline
{\color[HTML]{333333} B-V}                            & {\color[HTML]{333333} \textbf{0.32}} & {\color[HTML]{333333} \textbf{0.34}} & {\color[HTML]{333333} \textbf{-2.930}} & {\color[HTML]{333333} \textbf{0.010}} & {\color[HTML]{333333} 0.34}          & {\color[HTML]{333333} 0.37}          & {\color[HTML]{333333} -2.024}          & {\color[HTML]{333333} 0.061}          \\
{\color[HTML]{333333} B-A}                            & {\color[HTML]{333333} \textbf{0.29}} & {\color[HTML]{333333} \textbf{0.31}} & {\color[HTML]{333333} \textbf{-5.406}} & {\color[HTML]{333333} \textbf{0.000}} & {\color[HTML]{333333} \textbf{0.30}} & {\color[HTML]{333333} \textbf{0.33}} & {\color[HTML]{333333} \textbf{-2.494}} & {\color[HTML]{333333} \textbf{0.025}} \\
{\color[HTML]{333333} B-M}                            & {\color[HTML]{333333} \textbf{0.34}} & {\color[HTML]{333333} \textbf{0.35}} & {\color[HTML]{333333} \textbf{-3.230}} & {\color[HTML]{333333} \textbf{0.006}} & {\color[HTML]{333333} \textbf{0.35}} & {\color[HTML]{333333} \textbf{0.38}} & {\color[HTML]{333333} \textbf{-2.190}} & {\color[HTML]{333333} \textbf{0.045}} \\
{\color[HTML]{333333} B-P}                            & {\color[HTML]{333333} \textbf{0.27}} & {\color[HTML]{333333} \textbf{0.29}} & {\color[HTML]{333333} \textbf{-4.881}} & {\color[HTML]{333333} \textbf{0.000}} & {\color[HTML]{333333} \textbf{0.28}} & {\color[HTML]{333333} \textbf{0.31}} & {\color[HTML]{333333} \textbf{-2.530}} & {\color[HTML]{333333} \textbf{0.023}} \\
{\color[HTML]{333333} B-S}                            & {\color[HTML]{333333} \textbf{0.30}} & {\color[HTML]{333333} \textbf{0.31}} & {\color[HTML]{333333} \textbf{-3.060}} & {\color[HTML]{333333} \textbf{0.008}} & {\color[HTML]{333333} 0.29}          & {\color[HTML]{333333} 0.32}          & {\color[HTML]{333333} -1.993}          & {\color[HTML]{333333} 0.065}          \\
{\color[HTML]{333333} V-A}                            & {\color[HTML]{333333} \textbf{0.25}} & {\color[HTML]{333333} \textbf{0.26}} & {\color[HTML]{333333} \textbf{-5.646}} & {\color[HTML]{333333} \textbf{0.000}} & {\color[HTML]{333333} \textbf{0.27}} & {\color[HTML]{333333} \textbf{0.30}} & {\color[HTML]{333333} \textbf{-3.443}} & {\color[HTML]{333333} \textbf{0.004}} \\
{\color[HTML]{333333} V-M}                            & {\color[HTML]{333333} 0.40}          & {\color[HTML]{333333} 0.42}          & {\color[HTML]{333333} -1.401}          & {\color[HTML]{333333} 0.182}          & {\color[HTML]{333333} 0.41}          & {\color[HTML]{333333} 0.44}          & {\color[HTML]{333333} -1.545}          & {\color[HTML]{333333} 0.143}          \\
{\color[HTML]{333333} V-P}                            & {\color[HTML]{333333} \textbf{0.23}} & {\color[HTML]{333333} \textbf{0.25}} & {\color[HTML]{333333} \textbf{-7.293}} & {\color[HTML]{333333} \textbf{0.000}} & {\color[HTML]{333333} \textbf{0.26}} & {\color[HTML]{333333} \textbf{0.30}} & {\color[HTML]{333333} \textbf{-3.589}} & {\color[HTML]{333333} \textbf{0.003}} \\
{\color[HTML]{333333} V-S}                            & {\color[HTML]{333333} 0.32}          & {\color[HTML]{333333} 0.34}          & {\color[HTML]{333333} -1.657}          & {\color[HTML]{333333} 0.118}          & {\color[HTML]{333333} 0.31}          & {\color[HTML]{333333} 0.34}          & {\color[HTML]{333333} -1.357}          & {\color[HTML]{333333} 0.195}          \\
{\color[HTML]{333333} A-M}                            & {\color[HTML]{333333} \textbf{0.27}} & {\color[HTML]{333333} \textbf{0.30}} & {\color[HTML]{333333} \textbf{-2.908}} & {\color[HTML]{333333} \textbf{0.011}} & {\color[HTML]{333333} 0.28}          & {\color[HTML]{333333} 0.31}          & {\color[HTML]{333333} -1.801}          & {\color[HTML]{333333} 0.092}          \\
{\color[HTML]{333333} A-P}                            & {\color[HTML]{333333} \textbf{0.31}} & {\color[HTML]{333333} \textbf{0.34}} & {\color[HTML]{333333} \textbf{-2.366}} & {\color[HTML]{333333} \textbf{0.032}} & {\color[HTML]{333333} 0.32}          & {\color[HTML]{333333} 0.35}          & {\color[HTML]{333333} -2.086}          & {\color[HTML]{333333} 0.054}          \\
{\color[HTML]{333333} A-S}                            & {\color[HTML]{333333} \textbf{0.24}} & {\color[HTML]{333333} \textbf{0.26}} & {\color[HTML]{333333} \textbf{-4.392}} & {\color[HTML]{333333} \textbf{0.001}} & {\color[HTML]{333333} 0.23}          & {\color[HTML]{333333} 0.26}          & {\color[HTML]{333333} -1.883}          & {\color[HTML]{333333} 0.079}          \\
{\color[HTML]{333333} M-P}                            & {\color[HTML]{333333} \textbf{0.33}} & {\color[HTML]{333333} \textbf{0.35}} & {\color[HTML]{333333} \textbf{-3.257}} & {\color[HTML]{333333} \textbf{0.005}} & {\color[HTML]{333333} \textbf{0.33}} & {\color[HTML]{333333} \textbf{0.36}} & {\color[HTML]{333333} \textbf{-2.323}} & {\color[HTML]{333333} \textbf{0.035}} \\
{\color[HTML]{333333} M-S}                            & {\color[HTML]{333333} 0.29}          & {\color[HTML]{333333} 0.30}          & {\color[HTML]{333333} -2.657}          & {\color[HTML]{333333} 0.018}          & {\color[HTML]{333333} 0.31}          & {\color[HTML]{333333} 0.33}          & {\color[HTML]{333333} -2.919}          & {\color[HTML]{333333} 0.011}          \\
{\color[HTML]{333333} P-S}                            & {\color[HTML]{333333} 0.25}          & {\color[HTML]{333333} 0.29}          & {\color[HTML]{333333} -2.387}          & {\color[HTML]{333333} 0.031}          & {\color[HTML]{333333} 0.23}          & {\color[HTML]{333333} 0.26}          & {\color[HTML]{333333} -2.390}          & {\color[HTML]{333333} 0.030}          \\ \hline
\end{tabular}
\label{tab3}
\end{center}
\footnotesize{$^*$IS: imagined speech; VI: visual imagery; B:  Broca and Wernicke’s areas; V: visual cortex; A: auditory cortex; M: motor cortex; P: prefrontal cortex; S: sensory cortex}
\end{table}

\section{RESULTS AND DISCUSSION}

\subsection{PLV Analysis}

The PLV analysis for the imagined speech and visual imagery paradigms is presented in Tables \ref{tab1} and \ref{tab2}, revealing distinct neural synchronization patterns across tasks and subjects. 

In the imagined speech paradigm (Table \ref{tab1}), average PLV values ranged from 0.27 to 0.29, indicating moderate neural phase consistency across all classes. Among the classes, \textit{"Thank you"} and \textit{"Help me"} exhibited slightly higher PLV values, suggesting stronger phase-locking activity during these tasks. The standard deviation values, ranging from 0.04 to 0.11, highlight inter-subject variability in synchronization strength. Notably, subject S2 demonstrated elevated PLV values for multiple classes, indicating heightened neural engagement during imagined speech tasks. These findings underscore the utility of imagined speech as a reliable paradigm for engaging language-related neural networks.

In the visual imagery paradigm (Table \ref{tab2}), the average PLV values were slightly higher, ranging from 0.28 to 0.30. This suggests that visual imagery tasks elicited more robust neural phase alignment compared to imagined speech. Classes such as \textit{"Hello"} and \textit{"Light"} displayed comparatively higher PLV values, reflecting relatively strong neural synchronization during these tasks. The standard deviation values ranged from 0.07 to 0.11, highlighting inter-subject variability. Subject S7 demonstrated significantly higher PLV values across several classes, likely due to heightened neural engagement during visual imagery tasks, which may reflect individual differences in task-specific cognitive strategies.

\subsection{Comparison of Paradigms}

The comparison between paradigms reveals that visual imagery generally produces higher PLV values, potentially due to its reliance on well-structured spatial and visual networks. In contrast, imagined speech exhibited more consistent synchronization patterns across subjects, emphasizing its suitability for language-driven tasks. These differences highlight the distinct neural mechanisms underlying the two paradigms: visual imagery engages regions associated with spatial reasoning and visual processing, while imagined speech relies more heavily on language and sensory integration networks.



\subsection{Brain Region Analysis}

Table~\ref{tab3} shows brain connectivity patterns across cortical regions for imagined speech and visual imagery paradigms. Compared to the resting state, both paradigms displayed notable connectivity increases, particularly in Broca and Wernicke's areas, the auditory cortex, and the prefrontal cortex. For imagined speech, there was significant connectivity between Broca's area and both the auditory and prefrontal cortices, indicating strong engagement of language-related and cognitive processing networks, aligning with previous findings on imagined verbal communication \cite{lee2020neural}. In visual imagery, increased connectivity was observed between the visual cortex and both the prefrontal cortex and motor cortex, reflecting the activation of spatial and visual processing networks. While the auditory and motor cortices showed connectivity with other regions in both paradigms, the effect was more pronounced in imagined speech, possibly due to cross-modal activation in language tasks.


\subsection{Implications for BCI Applications}

From an application perspective, these findings suggest that imagined speech is well-suited for brain-computer interface (BCI) systems focused on word-based communication tasks, offering consistent neural synchronization and robust engagement of language-related regions. Visual imagery may serve as a complementary paradigm, particularly for users with speech impairments or a preference for visual-spatial tasks. Furthermore, the observed inter-subject variability in PLV and connectivity patterns highlights the importance of personalized calibration in BCI systems to optimize performance across diverse user populations. Future research should explore hybrid BCI paradigms that integrate imagined speech and visual imagery to leverage the strengths of both approaches. By addressing individual differences in neural dynamics and task engagement, such systems could enhance adaptability and reliability, paving the way for broader adoption of BCIs in real-world applications.


\section{CONCLUSION}
This study highlights the potential of imagined speech and visual imagery as effective BCI paradigms for communication. By analyzing functional connectivity, we identified distinct neural synchronization patterns for both paradigms, with visual imagery engaging spatial networks and imagined speech activating language-related regions. The inter-subject variability observed underscores the importance of personalized calibration to optimize BCI performance. Overall, these results underscore the potential of both imagined speech and visual imagery as intuitive BCI paradigms, providing a foundation for developing more personalized and adaptive BCI communication systems. Future work should expand on this research by including more diverse word classes and advanced deep learning models to enhance feature decoding, paving the way for personalized and adaptable BCI communication systems \cite{zhang2022unsupervised, kim2023diff}.

\bibliographystyle{jabbrv_IEEEtran}
\bibliography{REFERENCE}





\end{document}